\RequirePackage{amsmath}
\documentclass[runningheads]{llncs}
\usepackage[T1]{fontenc}
\usepackage{amsfonts}
\usepackage{graphicx}

\begin{document}
\title{Joint Dense-Point Representation for Contour-Aware Graph Segmentation}
\titlerunning{Joint Dense-Point Representation for Contour-Aware Graph Segmentation}
%
\author{Kit Mills Bransby\textsuperscript{1}, Greg Slabaugh\textsuperscript{1}, Christos Bourantas\textsuperscript{1,2}, Qianni Zhang\textsuperscript{1}}
\authorrunning{K. M. Bransby et al.}
%
\institute{Queen Mary University of London, United Kingdom \and Department of Cardiology, Barts Health NHS Trust, London, United Kingdom  \\
\email{\{k.m.bransby, qianni.zhang\}@qmul.ac.uk}}
\maketitle              
\begin{abstract}

We present a novel methodology that combines graph and dense segmentation techniques by jointly learning both point and pixel contour representations, thereby leveraging the benefits of each approach. This addresses deficiencies in typical graph segmentation methods where misaligned objectives restrict the network from learning discriminative vertex and contour features. Our joint learning strategy allows for rich and diverse semantic features to be encoded, while alleviating common contour stability issues in dense-based approaches, where pixel-level objectives can lead to anatomically implausible topologies. In addition, we identify scenarios where correct predictions that fall on the contour boundary are penalised and address this with a novel hybrid contour distance loss. Our approach is validated on several Chest X-ray datasets, demonstrating clear improvements in segmentation stability and accuracy against a variety of dense- and point-based methods. Our source code is freely available at: \\www.github.com/kitbransby/Joint\_Graph\_Segmentation

\keywords{Semantic Segmentation  \and Graph Convolutional Networks}
\end{abstract}
\section{Introduction}

Semantic segmentation is a fundamental task in medical imaging used to delineate regions of interest, and has been applied extensively in diagnostic radiology. Recently, deep learning methods that use a dense probability map to classify each pixel such as UNet~\cite{unet}, R-CNN~\cite{r-cnn}, FCN~\cite{fcn} have advanced the state-of-the-art in this area. Despite overall excellent performance, dense-based approaches learn using a loss defined at the pixel-level which can lead to implausible segmentation boundaries such as unexpected interior holes or disconnected blobs~\cite{gaggion}. This is a particular problem in medical image analysis where information-poor, occluded or artefact-affected areas are common and often limit a network's ability to predict reasonable boundaries. Furthermore, minimising the largest error (Hausdorff distance (HD)) is often prioritised over general segmentation metrics such as Dice Similarity (DS) or Jaccard Coefficient (JC) in medical imaging, as stable and trustworthy predictions are more desirable. 
\\
\indent To address this problem in segmentation networks, Gaggion \emph{et al.} proposed HybridGNet \cite{gaggion} that replaces the convolutional decoder in UNet with a graph convolutional network (GCN), where images are segmented using a polygon generated from learned points. Due to the relational inductive bias of graph networks where features are shared between neighbouring nodes in the decoder, there is a natural smoothing effect in predictions leading to stable segmentation and vastly reduced HD. In addition this approach is robust to domain shift and can make reasonable predictions on unseen datasets sourced from different medical centres, whereas dense-based methods fail due to domain memorization \cite{gaggion_isbi}. In HybridGNet, improved stability and HD comes at the cost of reduced contour detail conveyed by sub-optimal DS and JC metrics when compared to dense-based approaches such as UNet. Many methods have addressed this problem by rasterizing polygon points predicted by a decoder to a dense mask and then training the network using typical pixel-level losses such as Dice or cross-entropy ~\cite{curvegcn,boundaryformer,ACDRNet}. These approaches have merit but are often limited by their computational requirements. For example, in CurveGCN \cite{curvegcn}, the rasterization process uses OpenGL polygon triangulation which is not differentiable, and the gradients need to be approximated using Taylor expansion which is computationally expensive and can therefore only be applied at the fine-tuning stage \cite{taylor}. While in ACDRNet \cite{ACDRNet}, rasterization is differentiable, however the triangulation process is applicable only to convex polygons, and therefore limits application to more complicated polygon shapes. Rasterization is extended to non-convex polygons in BoundaryFormer \cite{boundaryformer} by bypassing the triangulation step and instead approximating the unsigned distance field. This method gives excellent results on MS-COCO dataset~\cite{MS-COCO}, however is computationally expensive (see Section \ref{comparison}).
\\
\indent With this in mind, we return to HybridGNet which efficiently optimises points directly and theorise about the causes of the performance gap relative to dense segmentation models. We identify that describing segmentation contours using points is a sub-optimal approach because (1) points are an incomplete representation of the segmentation map; (2) the supervisory signal is usually weaker ($n$ distances are calculated from $n$ pairs of points, versus, $h$ x $w$ distances for pairs of dense probability maps); (3) the distance from the contour is more meaningful than the distance from the points representing the contour, hence minimising the point-wise distance can lead to predictions which fall on the contour being penalised. 
\\
\indent \emph{Contributions}: We propose a novel joint architecture and contour loss to address this problem that leverages the benefits of both point and dense approaches. First, we combine image features from an encoder trained using a point-wise distance with image features from a decoder trained using a pixel-level objective. Our motivation is that contrasting training strategies enable diverse image features to be encoded which are highly detailed, discriminative and semantically rich when combined. Our joint learning strategy benefits from the segmentation accuracy of dense-based approaches, but without topological errors that regularly afflict models trained using a pixel-level loss. 
Second, we propose a novel hybrid contour distance (HCD) loss which biases the distance field towards predictions that fall on the contour boundary using a sampled unsigned distance function which is fully differentiable and computationally efficient. To our knowledge this is the first time unsigned distance fields have been applied to graph segmentation tasks in this way. Our approach is able to generate highly plausible and accurate contour predictions with lower HD and higher DS/JC scores than a variety of dense and graph-based segmentation baselines. 

\section{Methods}

\subsection{Network Design}
\begin{figure}[h!]
\includegraphics[width=\textwidth]{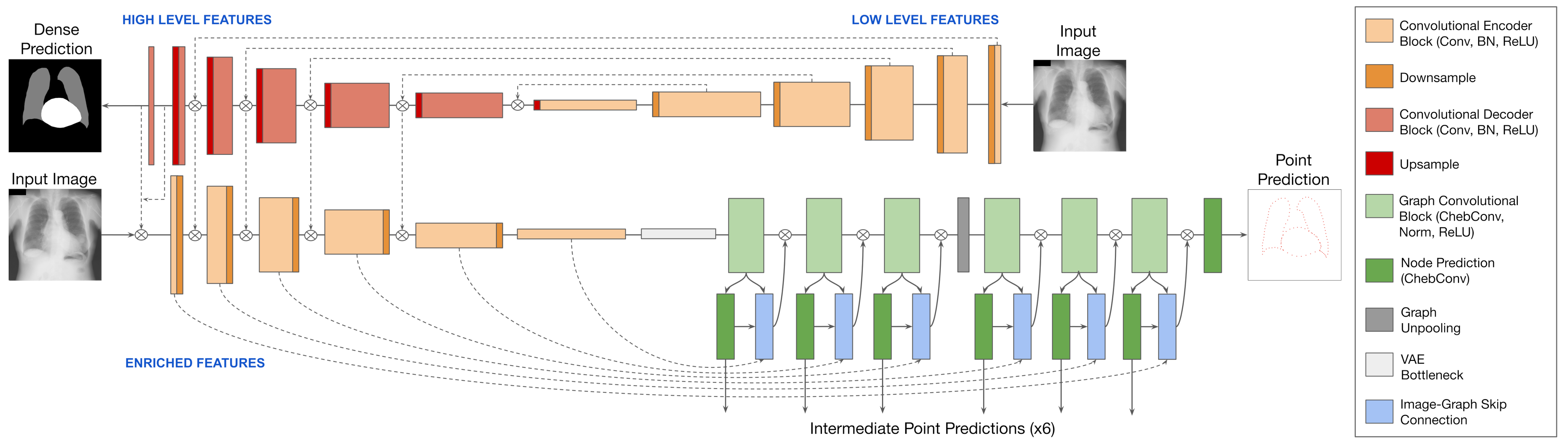}
\setlength{\belowcaptionskip}{-10pt}
\caption{Network Architecture: a Dense-Dense network (top) enriches image features in a Dense-Graph network (bottom). } 
\label{architecture}
\end{figure}

We implement an architecture consisting of two networks, a Dense-Graph (DG) network and a Dense-Dense (DD) network, as shown in Fig \ref{architecture}. Each network takes the same image input $X$ of height $H$ and width $W$ with skip connections passing information from the decoder of DD to the encoder of DG. For DG, we use a HybridGNet-style architecture containing a convolutional encoder to learn image features at multiple resolutions, and a graph convolutional decoder to regress the 2D coordinates of each point. In DG, node features are initialised in a variational autoencoder (VAE) bottleneck where the final convolutional output is flattened to a low dimensional latent space vector $z$. We sample $z$ from a distribution $Normal(\mu,\sigma)$ using the reparameterization trick \cite{reparameterization}, where $\mu$ and $\sigma$ are learnt parameters of the encoder. Image-to-Graph Skip Connections (IGSC) \cite{gaggion} are used to sample dense feature maps $F_{I} \in \mathbb{R}^{H \times W \times C}$ from DG's encoder using node position predictions $P \in \mathbb{R}^{N \times 2}$ from DG's graph decoder and concatenate these with previous node features $F_{G} \in \mathbb{R}^{N \times f}$ to give new node features $F_{G}^{\prime} \in \mathbb{R}^{N \times (f+C+2)}$. Here, $N$ is the number of nodes in the graph and $f$ is the dimension of the node embedding. We implement IGSC at every encoder-decoder level and pass node predictions as output, resulting in seven node predictions. For DD, we use a standard UNet using the same number of layers and dimensions as the DG encoder with a dense segmentation prediction at the final decoder layer. 

\subsection{Graph Convolutional Network}

Our graph decoder passes features initialised from the VAE bottleneck through six Chebyshev spectral graph convolutional \cite{chebconv} (ChebConv) layers using K-order polynomial filters. Briefly, this is defined by $X^{\prime} = \sigma(\sum_{K=1}^{k} Z^{(k)}\cdot \Theta^{(k)})$ where $\Theta^{(k)} \in \mathbb{R}^{f_{in} \: \times \: f_{out}}$ are learnable weights and $\sigma$ is a ReLU activation function. $Z^{(k)}$ is computed recursively such that $Z^{(1)} = X$ , $Z^{(2)} = \hat{L}\cdot Z^{(1)}$, $Z^{(k)} = 2\cdot \hat{L}\cdot Z^{(k-1)} - Z^{(k-2)}$ where $X \in \mathbb{R}^{N \: \times \: f_{in}}$ are graph features, and  $\hat{L}$ represents the scaled and normalized graph Laplacian \cite{pytorch-geometric}. In practice, this allows for node features to be aggregated within a K-hop neighbourhood, eventually regressing the 2D location of each node using additional ChebConv prediction layers ($f_{out} = 2$). As in \cite{gaggion}, our graph network also includes an unpooling layer after ChebConv block 3 to upsample the number of points by adding a new point in between existing ones.

\begin{figure}[h!]
\includegraphics[width=\textwidth]{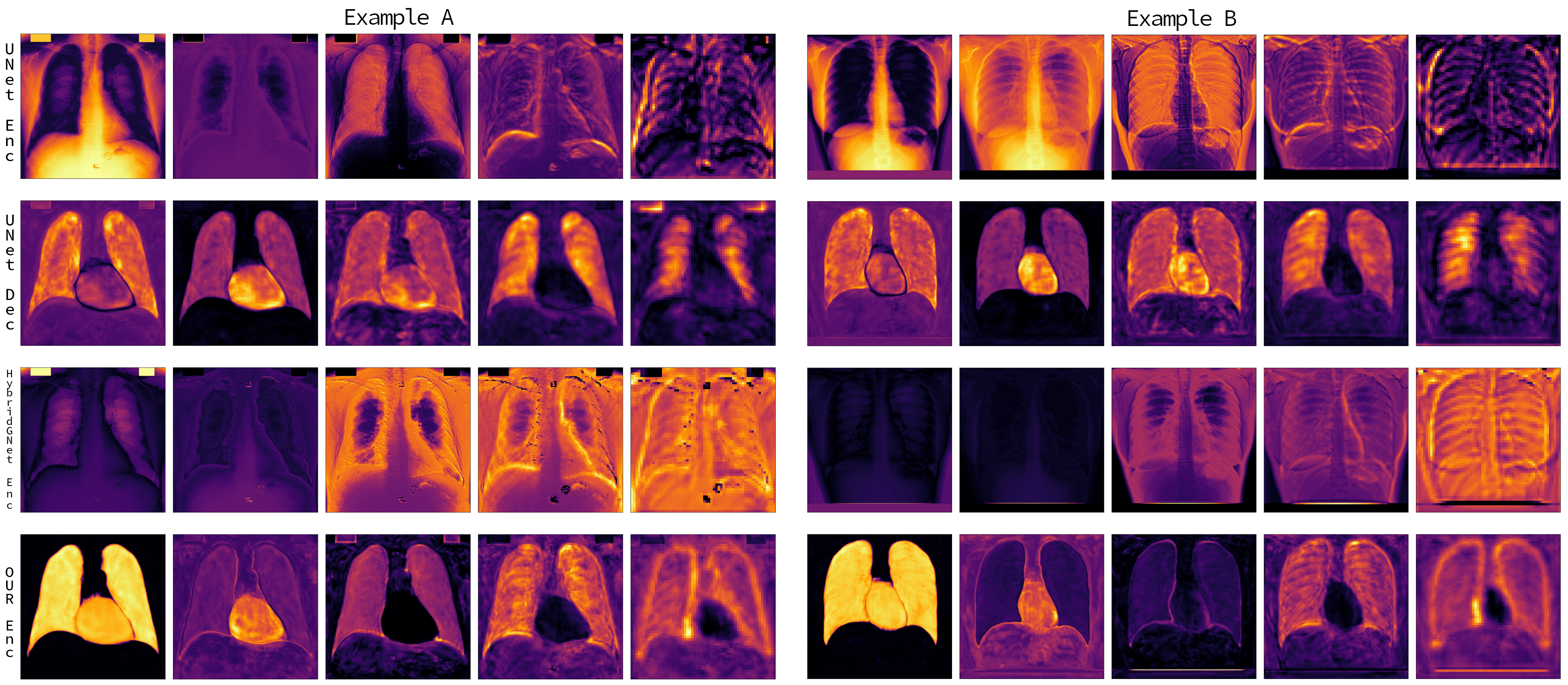}
\caption{Feature map activation comparison between UNet encoder, UNet decoder, HybridGNet encoder and our encoder, using two examples. Top four most activated channels are summed channel-wise for convolutional layers 1-5 in each encoder/decoder. L$\rightarrow$R: decreasing resolution, increasing channel depth. Note, activations in our encoder consistently highlight areas which are more pertinent to segmentation} \label{features}
\end{figure}

\subsection{Joint Dense-Point Learning}

As typical DG networks are trained with a point-wise distance loss and not a pixel-level loss, the image encoder is not directly optimised to learn clear and well-defined boundary features. This misalignment problem results in the DG encoder learning features pertinent to segmentation which are distinctively different from those learnt in DD encoders. This is characterised by activation peaks in different image regions such as the background and other non-boundary areas (see Fig \ref{features}). To leverage this observation, we enrich the DG encoder feature maps at multiple scales by fusing them with image features learnt by a DD decoder using a pixel-level loss. These diverse and highly discriminative features are concatenated before being passed through the convolutional block at each level. Current GCN feature learning paradigms aim at combining feature maps from neighbouring or adjacent levels so as to aggregate similar information. This results in a "coarse-to-fine" approach by first passing high level features to early graph decoder blocks, followed by low level features to late graph decoder blocks. Our joint learning approach is similar to this strategy but also supplements each DG encoder level with both semantically rich and highly detailed contour features learnt by the DD network. 

\begin{figure}[h!]
\centering
\includegraphics[width=1\textwidth]{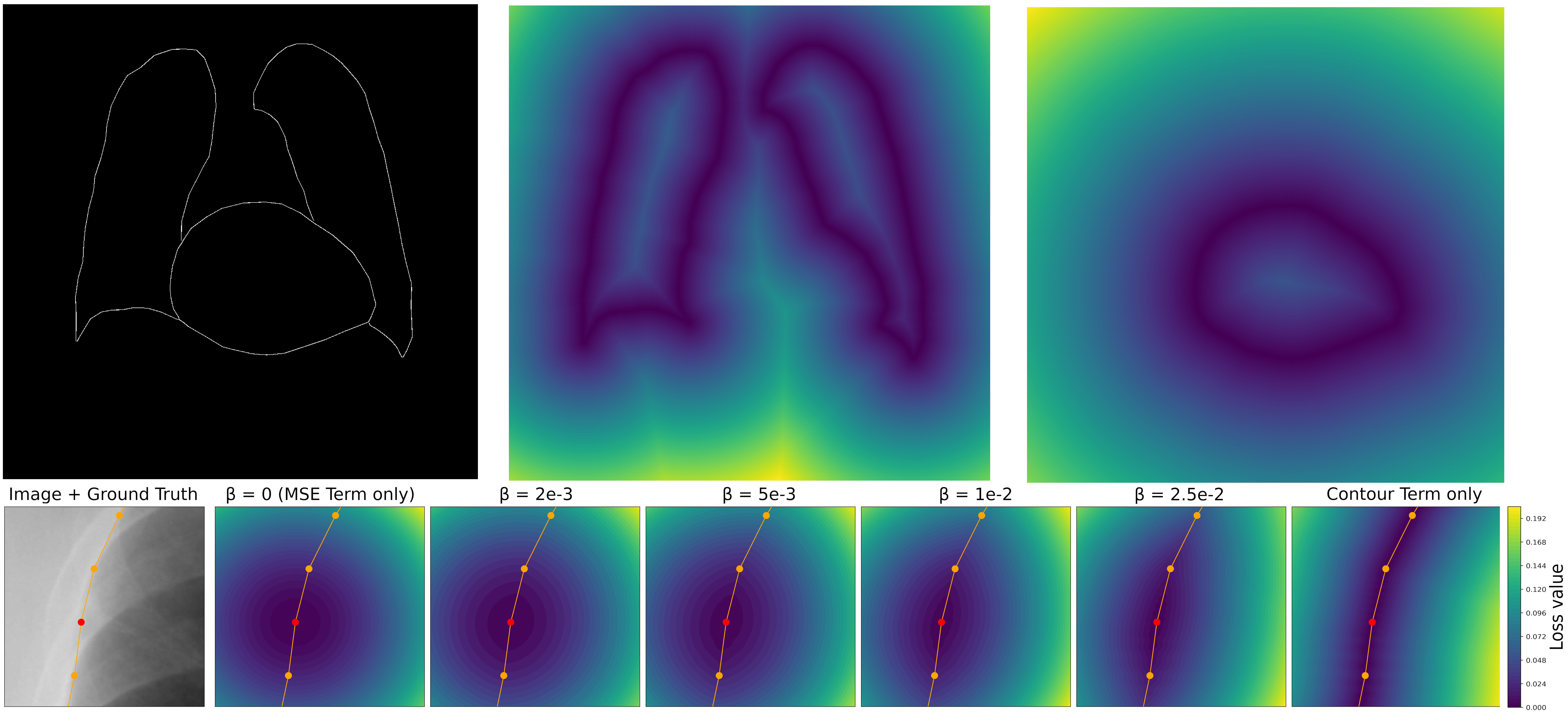}
\caption{Our Hybrid Contour Distance loss biases the distance field to contours rather than the points representing the contour. Top L$\rightarrow$R: Segmentation mask represented with edges, unsigned distance field for lungs, and heart. Bottom: Effect of beta in HCD.} \label{acad}
\end{figure}

\subsection{Hybrid Contour Distance}

Mean squared error (MSE) is a spatially symmetric loss which is agnostic to true contour borders. We alleviate this pitfall by designing an additional contour-aware loss term that is sensitive to the border. To achieve this we precompute a 2D unsigned distance map $S$ from the dense segmentation map for each class $c$ (i.e lungs, heart), where each position represents the normalised distance to the closest contour border of that class. Specifically, for a dense segmentation map $M$ we use a Canny filter~\cite{canny} to find the contour boundary $\delta M$ and then determine the minimum distance between a point $x \in c$ and any point $p$ on the boundary $\delta M_{c}$. This function is positive for both the interior and exterior regions, and zero on the boundary. Our method is visualised in Fig~\ref{acad} (first row) and formalised below:

\begin{equation}
S_{c}(\vec{x}) = \textup{min}\left | \vec{x} - \vec{p}\:  \right | \: \textup{for all} \: \: \vec{p} \in \delta M_{c}
\end{equation}

During training, we sample $S_{c}$ as an additional supervisory signal using the predicted 2D point coordinates $\hat{y}_{i} \in c$, and combine with MSE with weight $\beta$. The effect of $\beta$ is illustrated in Fig~\ref{acad} (second row) and full HCD loss function is defined below, where $N$ is the number of points and $y_{i} \in c$ is the ground truth point coordinate.

\begin{equation}
\mathcal{L}_{HCD} = \frac{1}{N} \sum_{i=1}^{N} [(y_{i} - \hat{y}_{i})^{2} + \beta S_{c}(\hat{y}_{i})]
\end{equation}

\section{Experiments and Results}\label{experiments}

\subsection{Datasets}

We obtain four publicly available Chest X-ray segmentation datasets (JSRT~\cite{JSRT}, Padchest~\cite{Padchest}, Montgomery~\cite{Montgomery}, and Shenzen~\cite{Shenzen}), with 245, 137, 566 and 138 examples respectively. JSRT cases are from patients diagnosed with lung nodules, while Padchest contains patients with a cardiomegaly diagnosis and features 20 examples where a pacemaker occludes the lung border. These two datasets contain heart and lung contour ground truth labels and are combined in a single dataset of 382 examples. Montgomery and Shenzen contain lung contour ground truth labels only, and are combined into a second dataset of 704 cases where 394 examples are from patients with tuberculosis and 310 are from patients without. Each combined dataset is randomly split into 70\% train, 15\% validation and 15\% test examples, each with a 1024px x 1024px resolution X-ray image and ground truth point coordinates for organ contours obtained from \cite{gaggion_isbi}. 

\subsection{Model Implementation \& Training}
We implement our model in PyTorch and use PyTorch-Geometric for the graph layer. All models were trained for 2500 epochs using a NVIDIA A100 GPU from Queen Mary's Andrena HPC facility. For reliable performance estimates, all models and baselines were trained from scratch three times, the mean scores obtained for quantitative analysis and the median model used for qualitative analysis. Hyperparameters for all experiments were unchanged from \cite{gaggion}. To impose a unit Gaussian prior on the VAE bottleneck we train the network with an additional KL-divergence loss term with weight $1e^{-5}$, and use $\beta = 2.5e^{-2}$ for the HCD weight. For joint models we pretrain the first UNet model separately using the recipe from \cite{gaggion} and freeze its weights when training the full model. This is done to reduce complexity in our training procedure. 

\begin{figure}[h!]
\includegraphics[width=\textwidth]{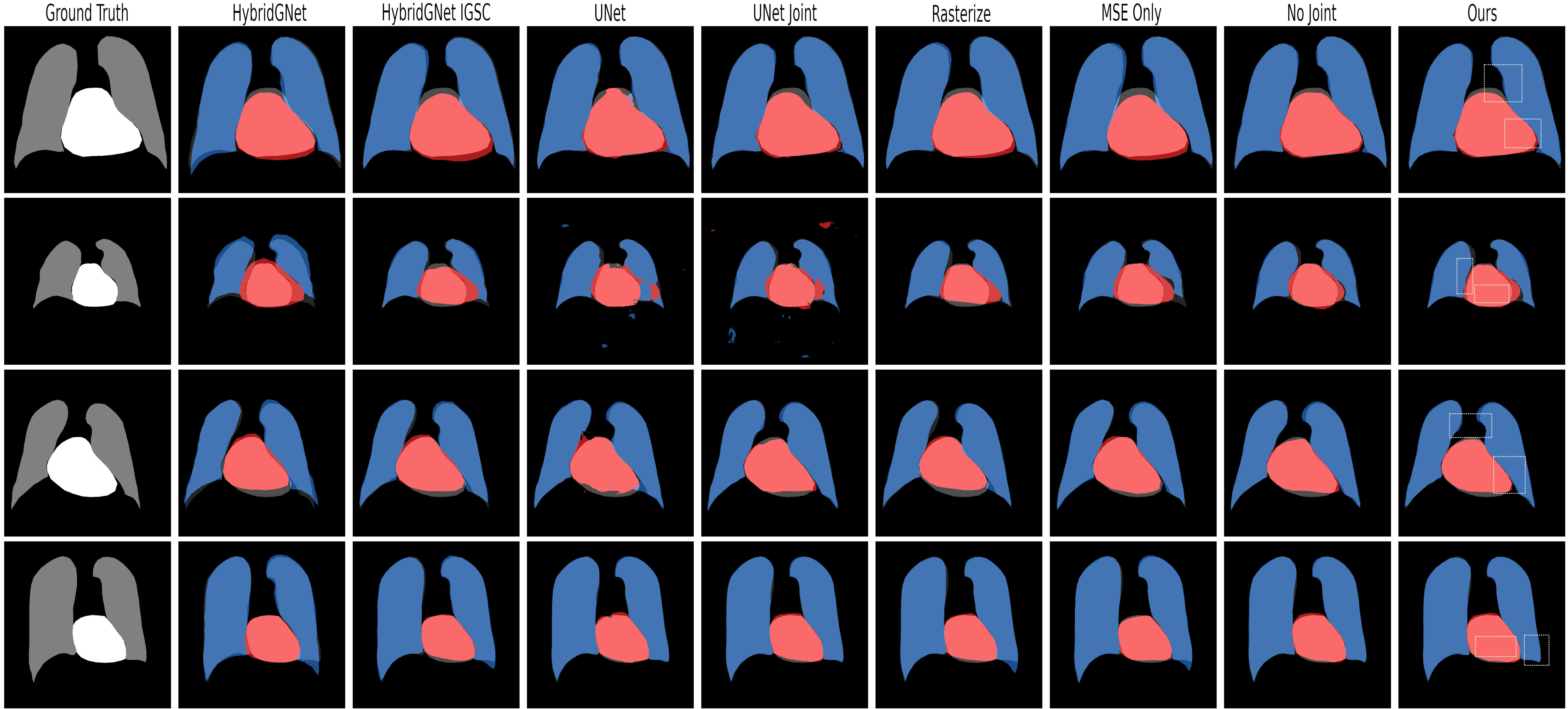}
\caption{JSRT \& Padchest: Qualitative Analysis. Note that our method does not suffer from the topological errors of dense-based methods but benefits from their segmentation accuracy. Specifically, improvements (white boxes) are most prevalent in areas of complexity such as where the heart and lungs intersect.} \label{qual}
\end{figure}

\subsection{Comparison to Existing Methods \& Ablation Study} \label{comparison}

We compare our approach to a variety of different dense- and point-based segmentation methods. First we validate our joint DD-DG learning approach by comparing to a DD-only segmentation network (UNet \cite{unet}) and DG-only segmentation networks (HybridGNet \cite{gaggion_miccai}, HybridGNet+ISGC \cite{gaggion}).

Next, we explore five alternative configurations of our joint architecture to demonstrate that our design choices are superior. These are: (1) UNet Joint: a network that uses our joint learning strategy but with two DD (UNet) networks, (2) Hourglass: joint learning but with no sharing between DD decoder and DG encoder, only the output of DD is passed to the input of DG, similar to the stacked hourglass network \cite{hourglass_2016,hourglass_2021}, (3) Hourglass Concat: as above, but the output of DD is concatenated with the input and both are passed to DG, (4) Multi-task: a single dense encoder is shared between a dense and graph decoder, similar to \cite{multi-task}, (5) No Joint: our network with no joint learning strategy. 

To demonstrate the effectiveness of our HCD loss, we compare to our joint network trained with the contour term removed (MSE only). Our HCD loss is similar to differentiable polygon rasterization in BoundaryFormer \cite{boundaryformer}, as they both use the distance field to represent points with respect to the true boundary. However, our method precomputes the distance field for each example and samples it during training, while BoundaryFormer approximates it on the fly. Hence we also compare to a single DG network (HybridGNet+IGSC) where each point output is rendered to a dense 1028px x 1028px segmentation map using rasterization and the full model is trained using a pixel-level loss. 

Table~\ref{jsrt_padchest_quant}-\ref{montgomery_shenzen_quant} demonstrate that our methodology outperforms all point- and dense-based segmentation baselines on both datasets. As seen in Fig~\ref{qual}, the performance increase from networks that combine image features from dense and point trained networks (column 7,9) is superior to when image features from two dense trained networks are combined (column 5). Furthermore, concatenating features at each encoder-decoder level (Table~\ref{jsrt_padchest_quant}-\ref{montgomery_shenzen_quant}, row 11) instead of at the input-output level (row 5-6) shows improved performance. The addition of HCD supervision to a DG model (Table~\ref{jsrt_padchest_quant}-\ref{montgomery_shenzen_quant}, row 8) gives similar improvements in segmentation when compared to using a differentiable rasterization pipeline (row 10), yet is far more computationally efficient (Table~\ref{montgomery_shenzen_quant}, column 7). 

\renewcommand{\arraystretch}{1.1}
\newsavebox\CBox
\def\textBF#1{\sbox\CBox{#1}\resizebox{\wd\CBox}{\ht\CBox}{\textbf{#1}}}

\begin{table}[]
\scriptsize
\centering
\begin{tabular}{p{26mm}p{11mm}p{15mm}p{10mm}p{10mm}p{10mm}p{10mm}p{10mm}p{10mm}}
\hline

                  & Predict & Supervision & \multicolumn{3}{l}{Lungs} & \multicolumn{3}{l}{Heart}  \\
                   &  &  & DC$\uparrow$ & HD$\downarrow$  & JC$\uparrow$ & DC$\uparrow$  & HD $\downarrow$      & JC$\uparrow$    \\  \hline 
HybridGNet        & point & point & 0.9313 & 17.0445 & 0.8731 & 0.9065 & 15.3786 & 0.8319 \\ 
HybridGNet+IGSC   & point & point & 0.9589 & 13.9955 & 0.9218 & 0.9295 & 13.2500 & 0.8702 \\
UNet              & dense & dense & 0.9665 & 28.7316 & 0.9368 & 0.9358 & 29.6317 & 0.8811 \\
UNet Joint        & dense & dense & 0.9681	& 26.3758 & 0.9395 & 0.9414	& 24.9409 & 0.8909 \\ 
Hourglass         & point & both & 0.9669 & 13.4225 & 0.9374 & 0.9441 & 12.3434 & 0.8954 \\ 
Hourglass Concat  & point & both & 0.9669 & 13.5275 & 0.9374 & 0.9438 & 12.1554 & 0.8948 \\ 
Multi-task        & point & both & 0.9610 & 15.0490	& 0.9257 & 0.9284 & 13.1997 & 0.8679 \\ 
No Joint          & point & point & 0.9655 & 13.2137 & 0.9341 & 0.9321 & 13.1826 & 0.8748 \\
MSE Only          & point & both & 0.9686 &  \textBF{12.4058} & 0.9402 & 0.9439 & 12.0872 & 0.8953 \\
Rasterize         & point & dense & 0.9659 & 13.7267 & 0.9349 & 0.9344 & 12.9118 & 0.8785 \\ 
Ours              & point & both &  \textBF{0.9698} & 13.2087 & \textBF{0.9423} & \textBF{0.9451} & \textBF{11.7721} & \textBF{0.8975}  \\ \hline
                  &       &       &         &        &        &         &        &     \\ [-2ex]  
\end{tabular}
\caption{JSRT \& Padchest Dataset: Quantitative Analysis} \label{jsrt_padchest_quant}
\end{table}

\begin{table}[]
\scriptsize
\centering
\begin{tabular}{p{28mm} p{12mm}p{15mm} p{12mm}p{12mm}p{12mm}p{18mm}}
\hline
                  & Predict & Supervision & DC$\uparrow$ & HD$\downarrow$  & JC$\uparrow$ & Inference (s)  \\ \hline
HybridGNet        & point & point & 0.9459 & 12.0294 & 0.8989 & 0.0433 \\
HybridGNet + IGSC & point & point & 0.9677 & 9.7591 & 0.9380 & 0.0448 \\
UNet              & dense & dense & 0.9716 & 16.7093 & 0.9453 & 0.0047 \\
UNet Joint        & dense & dense & 0.9713 & 16.5447 & 0.9447 & 0.0103 \\ 
Hourglass         & point & both & 0.9701 & 10.9284 & 0.9434 & 0.1213 \\ 
Hourglass Concat  & point & both & 0.9712 & 10.8193 & 0.9448 & 0.1218 \\ 
Multi-task        & point & both & 0.9697 & 10.8615 & 0.9417 & 0.0535 \\ 
No Joint          & point & point & 0.9701 & 9.8246 & 0.9424 & 0.0510 \\
MSE Only          & point & both & 0.9729 & 9.6527 & 0.9474 & 0.1224 \\ 
Rasterize         & point & dense & 0.9718 & \textBF{9.4485} & 0.9453 & 0.2421 \\ 
Ours              & point & both & \textBF{0.9732} & 10.2166 & \textBF{0.9481} & 0.1226 \\  \hline 
                  &       &       &         &        &    &    \\ [-2ex] 
\end{tabular}
\caption{Montgomery \& Shenzen Dataset: Quantitative Analysis + Inference Time} \label{montgomery_shenzen_quant}
\end{table}

\section{Conclusion}

We proposed a novel segmentation architecture which leverage the benefits of both dense- and point- based algorithms to improve accuracy while reducing topological errors. Extensive experiments support our hypothesis that networks that utilise joint dense-point representations can encode more discriminative features which are both semantically rich and highly detailed. Limitations in segmentation methods using a point-wise distance were identified, and remedied with a new contour-aware loss function that offers an efficient alternative to differentiable rasterization methods. Our methodology can be applied to any graph segmentation network with a convolutional encoder that is optimised using a point-wise loss, and our experiments across four datasets demonstrate that our approach is generalizable to new data. 

\subsubsection{Acknowledgements} This research is part of AI-based Cardiac Image Computing (AICIC) funded by the faculty of Science and Engineering at Queen Mary University of London. 

%
%
%
%

\end{document}